\documentclass{article}
\usepackage[dvipsnames]{xcolor}

\usepackage[dvipsnames]{xcolor}

\usepackage{url}

\usepackage{algorithm}
\usepackage{algorithmic}

\usepackage[utf8]{inputenc} 
\usepackage[T1]{fontenc}    
\usepackage{multirow}
\usepackage{tikz}

\usepackage{booktabs}
\usepackage[font=small]{caption}

\usepackage{url}            
\usepackage{booktabs}       
\usepackage{amsfonts}       
\usepackage{nicefrac}       
\usepackage{microtype}      
\usepackage{xcolor}         
\usepackage{amsmath}

\usepackage[preprint, nonatbib]{neurips_2025}

\usepackage{amsmath, amssymb, amsfonts}
\usepackage{bbm}
\usepackage{dsfont}
\usepackage{hyperref}
\usepackage{graphicx}
\usepackage{booktabs}
\usepackage{multirow}
\usepackage{caption}
\usepackage{pifont}
\usepackage{float}
\usepackage{subcaption}

\title{UnCLe: Benchmarking \underline{Un}supervised \underline{C}ontinual \underline{Le}arning for Depth Completion}

\author{
Xien Chen, Rit Gangopadhyay, Michael Chu, Patrick Rim, Hyoungseob Park, Alex Wong
}

\renewcommand\footnotemark{}

\begin{document}

\maketitle

\begin{abstract}
We propose UnCLe, the first standardized benchmark for \textbf{Un}supervised \textbf{C}ontinual \textbf{Le}arning of a multimodal 3D reconstruction task: Depth completion aims to infer a dense depth map from a pair of synchronized RGB image and sparse depth map. We benchmark depth completion models under the practical scenario of unsupervised learning over continuous streams of data. While unsupervised learning of depth boasts the possibility to continuously learn novel data distributions over time, existing methods are typically trained on a static, or stationary, dataset. However, when adapting to novel nonstationary distributions, they ``catastrophically forget'' previously learned information. UnCLe simulates these non-stationary distributions by adapting depth completion models to sequences of datasets containing diverse scenes captured from distinct domains using different visual and range sensors. We adopt representative methods from continual learning paradigms and translate them to enable unsupervised continual learning of depth completion. We benchmark these models across indoor and outdoor environments, and investigate the degree of catastrophic forgetting through standard quantitative metrics. We find that unsupervised continual learning of depth completion is an open problem, and we invite researchers to leverage UnCLe as a foundational development platform.
\end{abstract}


\section{Introduction}
Autonomous navigation, robotic manipulation, and augmented, virtual, and extended realities (AR/VR/XR) are some of the many applications that rely on a reconstruction of the three-dimensional (3D) environment or scene.
These applications are deployed on systems with heterogeneous sensors. Amongst them, visual (e.g., RGB camera) and range (e.g., time-of-flight, lidar) sensors are vital for recovering a metric-scale reconstruction, in the form of an egocentric range or depth map, to support these spatial applications. Fusion of returns from these two sensors necessitates a multimodal model that takes synchronized calibrated RGB image and sparse point cloud (typically projected onto the image plane as a sparse depth map) as input to recover a dense depth map, i.e., depth completion.


Depth completion models can be trained in a supervised \cite{park2020non,kam2022costdcnet,tang2024bilateral} or unsupervised manner \cite{wong2020unsupervised,wong2021unsupervised,yan2023desnet}. The former relies on ground truth, which is prohibitively expensive to acquire; what is called "ground truth" is actually the result of offline data processing that involves aggregation of sequences of RGB images and sparse point clouds followed by manual annotation to clean up erroneous or ambiguous regions \cite{geiger2012we}, which is not scalable. The latter relies on principles of Structure-from-Motion (SfM) \cite{frahm2010building,schonberger2016structure}, which is realized as a joint optimization of depth and pose estimation by minimizing RGB image and sparse range reconstruction objectives. Without the need for human annotation, unsupervised depth completion models boast the potential to continually learn and adapt to new 3D environments from the virtually limitless amount of un-annotated data in a life-long fashion.

\begin{figure}[t]
    \centering
    \includegraphics[width=1\linewidth]{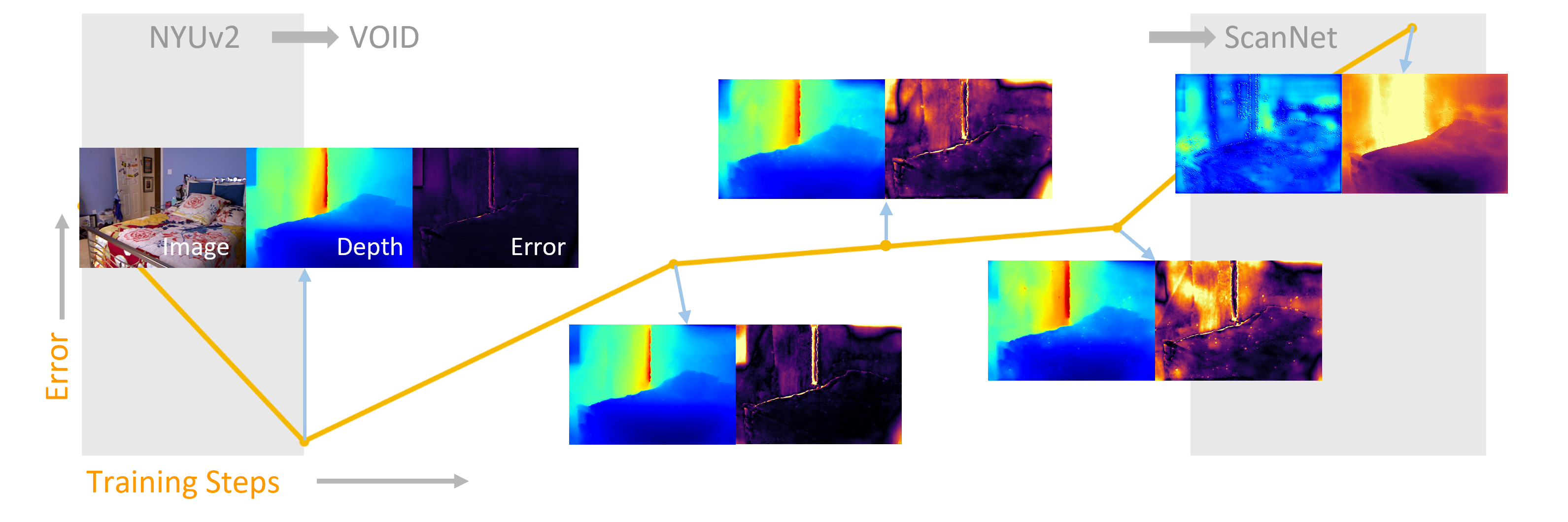}
    \vspace{-6mm}
    \caption{
      Error progression during continual unsupervised depth completion with a state-of-the-art continual learning method (ANCL). The model is initially on NYUv2, and continually trained on VOID and then ScanNet. Despite using a continual learning method, error increases over time, especially for later training steps. 
    }
    \label{fig:degrade}
    \vspace{-4mm}
\end{figure}

However, this potential is far from realized: existing depth completion models are predominately trained on a dataset under the assumption that the distribution of 3D scenes is stationary. When extended to a sequence of datasets, we find that as unsupervised depth completion models learn from or fit to new data, they inevitably "forget" previously learned information from past dataset(s). Due to evolving conditions and unseen domains, the updates to the model parameters lead to significant performance degradation on previously learned datasets when fitting to new data distributions \cite{french1999catastrophic,mccloskey1989catastrophic,thrun1995learning,ratcliff1990connectionist}. This phenomenon is commonly referred to as ``catastrophic forgetting.'' Even when applying mitigation strategies, we observe an increase in error as training progresses, as shown in Fig.~\ref{fig:degrade}, which underscores the difficulty of addressing forgetting in unsupervised depth completion.


One school of thought is to re-train these models on the new data along with all previously collected datasets offline. This amounts to increasing computational costs that scale with the size of newly collected data. Additionally, due to initialization and stochasticity of optimization, the newly trained model may exhibit errors absent in the old one \cite{yan2021positive} -- leading to performance regression. We, instead, subscribe to continual (or lifelong) learning, which is the training paradigm that addresses the challenge of catastrophic forgetting by enabling a single model to learn from a continuous stream of datasets, adapting to new datasets while preserving performance on previous ones. Therefore, we investigate unsupervised continual learning for depth completion, aiming to enable a multimodal depth estimation model to learn new datasets without forgetting, with the goal of fostering research that will culminate in models that can learn and adapt their inference without human intervention. 

\begin{figure}[t]
    \centering
    \includegraphics[width=0.92\linewidth]{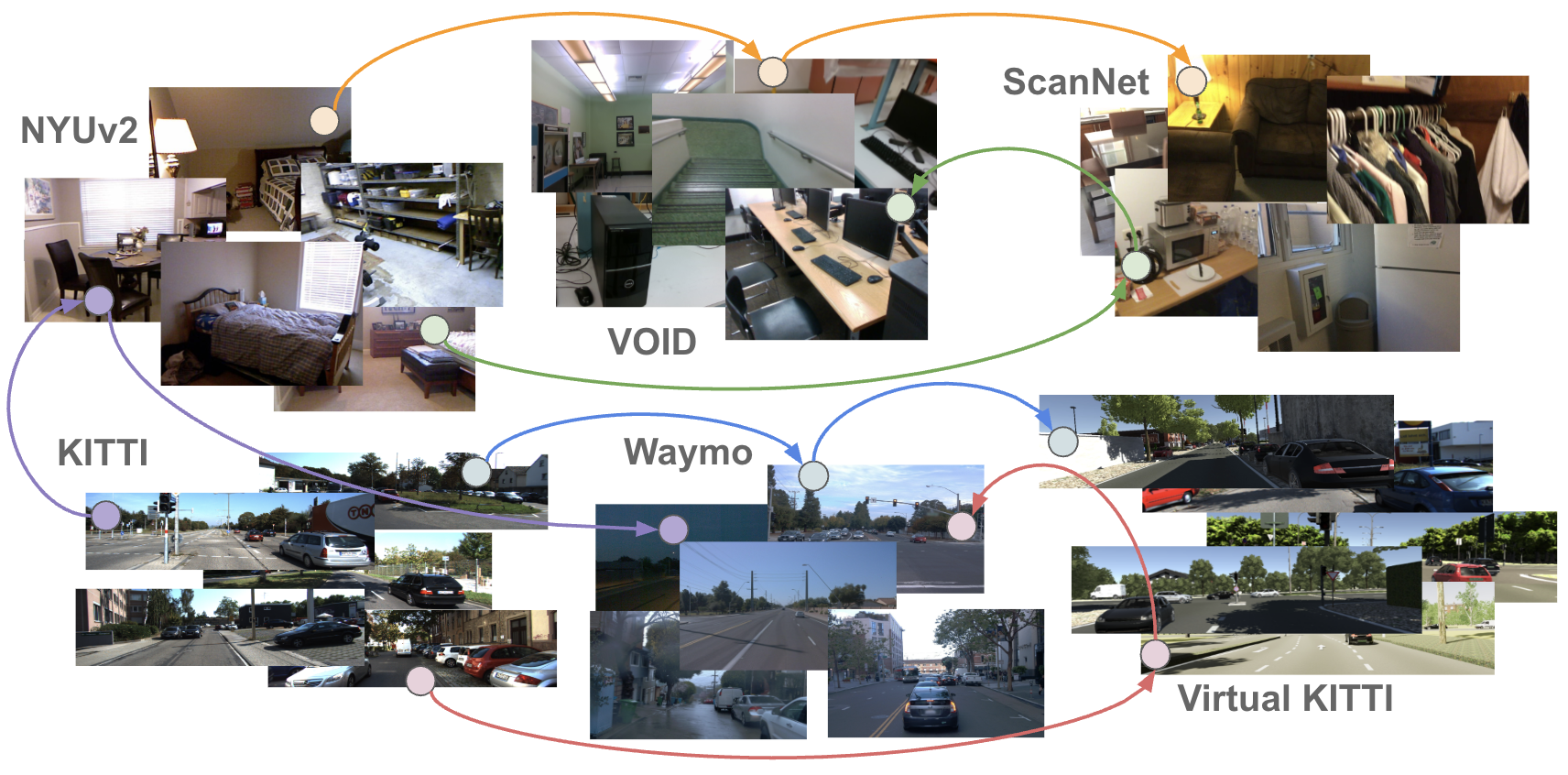}
    \vspace{-3mm}
    \caption{
        \textbf{Overview of benchmark.} UnCLe comprises six diverse datasets that are arranged into five sequences including indoor (orange, green arrows), outdoor (blue, red arrows), and indoor-outdoor (purple arrow). Each colored arrow path indicates a distinct dataset sequence used in our benchmark (e.g., NYUv2 $\rightarrow$ ScanNet $\rightarrow$ VOID for indoor, and KITTI $\rightarrow$ Waymo $\rightarrow$ Virtual KITTI for outdoor). Methods are evaluated on Average Forgetting, Average Performance, and Stability-Plasticity Trade-Off over four standard performance metrics.
    }
    \label{fig:path}
    \vspace{-2mm}
\end{figure}

To this end, we propose a comprehensive benchmark for unsupervised continual learning of depth completion, termed "UnCLe". We begin with the motivating observation of how the performance of three unsupervised depth completion models~\cite{wong2020unsupervised, wong2021unsupervised,wong2021learning} degrades after being finetuned on sequences of datasets. We adopt five canonical continual learning methods (i.e., regularization-based~\cite{kirkpatrick2017overcoming,li2017learning,kim2022ancl} and rehearsal-based~\cite{rolnick2019experience,kang2024continual} methods) and translate them for depth completion to mitigate catastrophic forgetting in this multimodal 3D reconstruction task. The benchmark evaluates methods on five dataset sequences, comprising six diverse datasets~\cite{wong2020unsupervised,geiger2012we, silberman2012indoor, dai2017scannet, sun2020scalability,gaidon2016virtual} that were collected with different cameras and range sensors (see Fig.~\ref{fig:path}).
UnCLe presents transitions between indoor scenes, outdoor scenes, and across indoor and outdoor scenes, a challenging scenario that no method has demonstrated performance on before. Through extensive hyper-parameter search for all depth completion and continual learning methods across all sequences, amounting to over a thousand experiments, we conclude that unsupervised continual learning for depth completion remains an open problem. UnCLe aims to serve as the first benchmark to streamline further research in this area.


Our contributions are as follows: (i) We propose UnCLe, the first standardized benchmark for unsupervised depth completion. (ii) We adapt existing, canonical regularization-based and rehearsal-based continual learning methods for unsupervised depth completion, presenting the first continual unsupervised depth completion methods. (iii) We introduce novel evaluation metrics tailored to continual depth completion—Average Forgetting, Average Performance, and Stability-Plasticity Trade-Off—to allow for comprehensive assessment of method behavior. (iv) Through 1,000+ experiments, we show that even the state-of-the-art continual learning methods exhibit a high degree of forgetting and performance for the task of depth completion on this challenging benchmark.


\section{Related Works}


\textbf{Regularization-based Continual Learning} approaches introduce constraints to the loss function to ensure that model behavior learned from previous datasets is not significantly altered. Amongst these methods, one research thread constrains changes to model parameters that are deemed important for previous datasets. Elastic Weight Consolidation (EWC) \cite{kirkpatrick2017overcoming} selectively decreases the plasticity of model weights determined by the Fisher information matrix. \cite{zenke2017continual} estimates parameter importance based on their contribution to loss changes during training.  \cite{aljundi2018memory} assesses importance by measuring the sensitivity of model outputs to parameter variations. \cite{chaudhry2018riemannian} combines \cite{kirkpatrick2017overcoming} and \cite{zenke2017continual} to leverage their strengths. Another thread focuses on function regularization, which aims to preserve model output behavior on previous datasets by constraining changes in intermediate features or final predictions through knowledge distillation (KD) \cite{hinton2015distilling}. Learning without Forgetting (LwF) \cite{li2017learning} leverages new data samples to approximate the responses of the old model. LwM \cite{dhar2019learning} integrates attention maps into the KD process to capture essential features, while EBLL \cite{rannen2017encoder} preserves feature reconstructions by utilizing dataset-specific autoencoders. GD \cite{lee2019overcoming} exploits external unlabeled data to extend regularization beyond the training set, while PODNet \cite{douillard2020podnet} and LUCIR \cite{hou2019learning} aim to preserve feature similarity. Bayesian approaches, such as FRCL \cite{titsias2019functional} and FROMP \cite{pan2020continual}, use probabilistic models to regularize the functional space. VCL \cite{nguyen2018variational} extends these ideas and uses variational inference to maintain a balance between stability and plasticity. More recently, ANCL~\cite{kim2022ancl} introduces an auxiliary network to mitigate forgetting through a learned regularizer. In this paper, we adopt EWC, LWF, and ANCL as they are milestone works of the regularization-based approaches.

\textbf{Rehearsal-based Continual Learning} approaches reintroduce previous observed examples during the training of new datasets. Experience Replay \cite{rolnick2019experience} uses a data ('replay') buffer to store actual samples from previous datasets to be trained on again to reinforce performance on previous datasets. Works along this thread propose data selection strategies for choosing the replay buffer, including randomized Reservoir Sampling \cite{chaudhry2019tiny, riemer2018learning, vitter1985random}, class-based sampling \cite{lopez2017gradient, rebuffi2017icarl}. Generative replay \cite{shin2017continual} utilizes GANs and VAEs to produce synthetic data mimicking previously observed data distributions. VAE-approaches \cite{riemer2019scalable, rostami2019complementary,pfulb2021continual,kemker2017fearnet,gopalakrishnan2022knowledge,ayub2021eec} are able to control the generated data labels, but suffer from blurry quality. GAN-approaches \cite{ayub2021eec,ostapenko2019learning,wu2018memory} improve the quality of the generated input data. Feature replay stores features instead of the raw data in order to reduce the storage burden. \cite{zhu2022self, liu2020generative, iscen2020memory} use feature distillation between new and old models. \cite{belouadah2019il2m} stores initial class statistics, like mean and covariance, to rectify biases in predictions. CMP~\cite{kang2024continual} improves memory efficiency and long-term adaptability by combining meta-representation learning with an optimal buffer replay strategy that selects diverse and representative samples based on representation similarity. UnCLe adopts Experience Replay and CMP as representative rehearsal-based continual learning methods. 


\textbf{Continual Learning for Depth Estimation.}
MonoDepthCL \cite{chawla2024continual} employs a dual-memory rehearsal-based method to address the challenges of catastrophic forgetting in unsupervised monocular depth estimation. CoDEPS \cite{vodisch2023codeps} employs a unique domain-mixing strategy for pseudo-label generation with efficient experience replay. However, previous works in this field typically focus on a single modality and lack standardized benchmarks, limiting their applicability to real-world scenarios where the use of multimodal data is standard. Our paper contributes to this field by establishing a benchmark for depth completion using both RGB images and sparse range data, addressing these gaps and setting a foundation for future research in continual depth estimation.


\textbf{Depth Completion}~\cite{yang2019dense,liu2022monitored,wu2024augundo,park2024test} is the task of inferring a dense metric depth from an RGB image and synchronized sparse point cloud.
\textit{Supervised methods} \cite{park2020non,hu2021penet,li2020multi,eldesokey2020uncertainty,van2019sparse,zhang2018deep,zhang2023completionformer,yu2023aggregating} minimizes discrepancy between model estimates and ground-truth dense depth maps. While effective, these approaches are limited by the availability and cost of obtaining ground truth, making them less practical for continual learning scenarios.
\textit{Unsupervised depth completion} methods \cite{wong2020unsupervised,yan2023desnet,wong2021learning,yang2019dense,wu2024augundo,ma2019self,shivakumar2019dfusenet,wong2021adaptive} are trained by minimizing a loss function comprising image and sparse depth reconstruction terms along with a local smoothness regularizer.
\cite{ma2019self} utilizes Perspective-n-Point \cite{lepetit2009epnp} and RANSAC \cite{fischler1981random} to align adjacent video frames.
\cite{yang2019dense} trains a depth prior conditioned on the image. FusionNet \cite{wong2021learning} leverages synthetic data to learn a prior on shapes, while \cite{lopez2020project} leverage sim2real adaptation to make use of rendered synthetic depth. VOICED \cite{wong2020unsupervised} approximates a scene with scaffolding.
\cite{wong2021adaptive} introduces an adaptive scheme to reduce penalties incurred on occluded regions. KBNet~\cite{wong2021unsupervised} proposes calibrated back-projection. \cite{yan2023desnet} decouples scale and structure.
\cite{jeon2022struct} uses visual SLAM features and \cite{liu2022monitored} proposes monitored distillation for positive congruent training.

However, all of these methods are subject to catastrophic forgetting as they rely on updating model weights to fit onto a new target dataset without the objective of retaining past information. This aspect makes it challenging for them to adapt to a continual learning setting, which aims to maintain performance on both new and previously seen datasets. We adopt VOICED, FusionNet, and KBNet and evaluate them over five representative continual learning methods. 
\section{Method Formulation} 
\textbf{Problem definition}. Unsupervised depth completion assumes a pair of synchronized RGB image and sparse depth map as input. Two time-adjacent RGB images are also assumed to be available during training. Let \( I : \Omega \subset \mathbb{R}^2 \rightarrow \mathbb{R}^3_+ \) denote an RGB image at current timestamp $t$ obtained from a calibrated camera, and \( z : \Omega \rightarrow \mathbb{R}_+ \) the corresponding sparse depth map derived from a projected 3D point cloud. Given the image \( I \), the sparse depth map \( z \), and the intrinsic calibration matrix \( K \in \mathbb{R}^{3 \times 3} \), the goal is to learn a function \( \hat{d} = f_\theta(I, z) \) that estimates a dense depth map. Unsupervised methods are trained by minimizing photometric reprojection error between the observed image $I_t$ and a reconstructed image $\hat{I}_t$ from adjacent frames $I_{\tau}$ for $\tau \in \{t-1, t+1\}$ using the estimated depth $\hat{d}$ and relative camera pose $g_{\tau t} \in SE(3)$:
\begin{equation}
\label{eqn:image_reconstruction}
    \hat{I}_t(x, \hat{d}, g_{\tau t}) = I_{\tau} \big( \pi  g_{\tau t} K^{-1} \bar{x} \hat d(x) \big).
\end{equation}
where $\pi$ denotes canonical perspective projection and \( \bar{x} \) the homogeneous coordinates \( [x^\top, 1]^\top \) of \( x \in \Omega \). $g_{\tau t}$ is given or estimated by a pose network and is only used during training.

In the continual learning setting, our objective is to adapt a pretrained model $f_\theta$, trained on an initial dataset $D_0 = \{( I_0^{(i)}, z_0^{(i)}, K_0^{(i)} ) \}_{i=1}^n$, to a sequence of new datasets $ D_1, D_2, \ldots, D_N $, where each $D_k = \{ ( I_{k}^{(i)}, z_{k}^{(i)}, K_{k}^{(i)} ) \}_{i=1}^{n_k}$ and $n_k$ denotes the number of training examples in $T_k$. The challenge is to incrementally train $f_\theta $ on each new target dataset $D_k$ without experiencing significant degradation in performance on the source dataset $S$ and all previously seen target datasets $D_{j}$ $\forall j < k$. This requires effectively mitigating catastrophic forgetting while enabling the model to adapt across different domains or datasets in the depth completion task.

\textbf{Unsupervised Depth Completion Loss}. We train \( f_\theta(I, z) \) by minimizing 
\begin{equation}
    \label{eqn:objective_function}
    \mathcal{L} = w_{ph}\ell_{ph}+w_{sz}\ell_{sz}+w_{sm}\ell_{sm}
\end{equation}
where each loss term can be weighted by their respective $w$ (see Sup. Mat. for details). The photometric consistency loss,
\begin{equation}
\ell_{ph} = \frac{1}{|\Omega|} \sum_{\tau \in T} \sum_{x \in \Omega} \left[ w_{co} \left| \hat{I}_{\tau}(x) - I(x) \right| + w_{st} \left( 1 - \texttt{SSIM}(\hat{I}_{\tau}(x), I(x)) \right) \right]
\label{eqn:photometric_consistency_loss}
\end{equation}
penalizes structural and color discrepancies between the reconstructed and observed images. The sparse depth consistency loss,
\begin{equation}
\label{eqn:sparse_depth_consistency_loss}
  	\ell_{sz} = \frac{1}{|\Omega|} 
  	    \sum_{x \in \Omega} 
  	       M(x) | \hat{d}(x) - z(x)| 
\end{equation}
grounds the predicted depth to metric scale, where $M : \Omega \subset \mathbb{R}^2 \rightarrow \{0, 1\}$ yields a binary mask of all valid sparse depth points. The local smoothness loss,
\begin{equation}
\label{eqn:local_smoothness_loss}
  	\ell_{sm} = \frac{1}{|\Omega|}
  	    \sum_{x \in \Omega} 
      	    \lambda_{X}(x)|\partial_{X}\hat{d}(x)|+
      	    \lambda_{Y}(x)|\partial_{Y}\hat{d}(x)|
\end{equation}
encourages smooth transitions of depth gradients in the $x-$ ($\partial_{X}$) and $y-$ ($\partial_{Y}$) directions, and is weighted by image gradients $\lambda_{X} = e^{-|\partial_{X}I_{t}(x)|}$ and $\lambda_{Y} = e^{-|\partial_{Y}I_{t}(x)|}$ to account for object edges.

\subsection{Elastic Weight Consolidation}
For the task of continual depth completion, we implement EWC as follows: When training on a new dataset $T_k$, we first load the previously trained model $f_{\theta^*}$, which is parameterized by the weight matrices $\theta^*$ optimized for $T_{k-1}$. These parameters are frozen and treated as a reference during training on $T_k$. The Fisher information matrix $F_i$ quantifies each parameter's $\theta_i$ importance to the previous model's performance on $T_{k-1}$. The EWC loss is therefore defined as:
\begin{equation}
\mathcal{L}_{\text{ewc}} =  \lambda_{ewc} \sum_i \frac{1}{2} F_i (\theta_i - \theta^*_{i})^2
\end{equation}
EWC loss penalizes important weights from moving too far away from their values optimized for the previous dataset. We add $\mathcal{L}_{\text{ewc}}$ to the unsupervised loss $\mathcal{L}$ (Eq. \ref{eqn:objective_function}) with $\lambda_{ewc} = 1$, which we found to be optimal after extensive tuning, when choosing EWC as the continual learning method.

\subsection{Learning without Forgetting}

To implement Learning without Forgetting (LwF) for the depth completion datasets, we aim to preserve output behavior from previous tasks while training only on new data. Specifically, we load the checkpoint of the old model \( f_{\theta^*} \), which is kept frozen during training. We compute a mean squared error (MSE) loss between the outputs of the frozen model and the new model \( f_\theta \). The LwF loss encourages the new model's predictions to remain consistent with that of the frozen model:
\begin{equation}
\mathcal{L}_{\text{lwf}} = \lambda_{lwf} \cdot ||f_{\theta}(I, z) - f_{\theta^*}(I, z)||_2^2
\end{equation}
where $\lambda_{lwf}=0.5$ is empirically determined. We add $\mathcal{L}_{\text{lwf}}$ to $\mathcal{L}$ (Eq. \ref{eqn:objective_function}) when choosing LwF as the continual learning method. LwF allows the model to learn from new depth data without requiring access to previous task data, as the frozen model's predictions act as a proxy for past knowledge.

\begin{table*}[t]
\scriptsize
\centering
\vspace{-2mm}
\caption{\textbf{Results on indoor.} Models are trained on NYUv2 and continually trained on ScanNet, then VOID. 
}
\setlength\tabcolsep{2pt}
\resizebox{1.0\textwidth}{!}{%
\begin{tabular}{c l c c c c@{\hspace{7pt}} c c c c@{\hspace{7pt}} c c c c}
    \toprule
    & & \multicolumn{4}{c}{Average Forgetting (\%)} & \multicolumn{4}{c}{Average Performance (mm)\hspace*{6pt}} & \multicolumn{4}{c}{SPTO (mm)} \\
    \midrule
    Model & Method & MAE & RMSE & iMAE & iRMSE & MAE & RMSE & iMAE & iRMSE & MAE & RMSE & iMAE & iRMSE \\ 
    \midrule
    \multirow{6}*{VOICED}
    & Finetuned
    & 8.828 & 6.131 & 6.951 & 7.042 & 63.352 & 125.28 & 15.461 & 35.053& 52.453 & 108.434 & 15.360 & 35.357 
    \\
    & EWC
    & 9.439 & 8.014 & 5.183 & 6.174 & 63.787 & 126.706 & 15.229 & 34.367 & 53.614 & 110.956 & 15.091 & 34.039
    \\
    & LwF
    & 8.591 & 8.456 & 9.613 & 21.774 & 65.135 & 126.968 & 16.221 & 38.002 & 53.517 & 108.845 & 15.402 & 34.729
    \\
    & Replay
    & 6.154 & 4.688 & 9.471 & 11.713 & 64.305 & 126.714 & 16.373 & 36.729 & 54.326 & 112.218 & 16.640 & 37.671
    \\
    & ANCL
    &6.796 & 5.870 & 6.206 & 14.586 & 86.653 & 134.761 & 167.249 & 203.559 & 80.039 & 139.843 & 91.895 & 121.101
    \\
    & CMP
    &-0.365 & -1.276 & 0.290 & 1.766 & 62.031 & 124.581 & 16.212 & 36.782 & 52.510 & 110.754 & 16.435 & 37.899
    \\
    \midrule
    \multirow{6}*{FusionNet} 
    & Finetuned
    & 24.928 & 9.775 & 32.333 & 16.799  & 66.523 & 130.142 & 15.829 & 33.881& 54.252 & 110.666 & 15.317 & 33.726
    \\
    & EWC
    & 11.256 & 8.782 & 17.944 & 17.847  & 64.487 & 130.890 & 15.264 & 34.203& 51.345 & 109.223 & 14.276 & 32.781
    \\
    & LwF
    & 6.863 & 2.865 & 7.336 & 1.939& 61.204 & 123.573 & 14.075 & 30.879 & 50.159 & 106.386 & 13.879 & 31.608 
    \\
    & Replay
    & 5.702 & 2.862 & 12.196 & 11.186  & 61.467 & 125.587 & 14.750 & 33.279& 50.273 & 108.608 & 14.351 & 33.658
    \\
    & ANCL
    & 3.952	&5.392	&8.765	&16.647 &68.963	&141.313 &18.380 &42.584 &57.390	&122.924 &18.044 &41.724
    \\
    & CMP
    & 2.516	&3.242	&4.208  &6.653 &69.633 &140.176	&18.303	&40.839 &57.642 &122.026 &18.234 &41.605 
    \\
    \midrule
    \multirow{6}*{KBNet}
    & Finetuned
    & 16.080 & 15.463 & 8.188 & 9.170 
    
    & 58.577 & 124.606 & 13.474 & 31.409& 47.890 & 105.807 & 13.266 & 31.742
    \\
    & EWC
    & 14.915 & 11.878 & 10.398 & 5.640
    & 57.414 & 122.075 & 13.741 & 31.552& 48.031 & 106.661 & 14.129 & 33.096
    \\
    & LwF
    & 9.717 & 6.324 & 6.168 & 5.254
    & 57.511 & 119.093 & 14.119 & 32.165& 47.154 & 103.164 & 14.304 & 33.838
    \\ 
    & Replay
    & 7.200 & 4.819 & 9.202 & 9.539
    & 56.208 & 117.848 & 13.983 & 32.341& 46.700 & 103.631 & 13.844 & 33.326
    \\
    & ANCL
    &9.73 & 10.75 & 5.58 & 16.38 & 56.89 & 120.30 & 13.77 & 31.85 & 47.32 & 103.42 & 13.88 & 32.76
    \\
    & CMP
    &5.39 & 5.11 & 8.25 & 7.90 & 55.92 & 117.83 & 13.74 & 31.43  & 46.03 & 102.36 & 13.55 & 32.03
    \\
    \midrule
\end{tabular}
}
\vspace{-3mm}
\label{tab:indoor1}
\end{table*}

\begin{table*}[t]
\scriptsize
\centering
\vspace{-2mm}
\caption{\textbf{Results on indoor.} Models are trained on NYUv2 and continually trained on VOID, then ScanNet. 
}
\setlength\tabcolsep{2pt}
\resizebox{1.0\textwidth}{!}{%
\begin{tabular}{c l c c c c@{\hspace{7pt}} c c c c@{\hspace{7pt}} c c c c}
    \toprule
    & & \multicolumn{4}{c}{Average Forgetting (\%)} & \multicolumn{4}{c}{Average Performance (mm)\hspace*{6pt}} & \multicolumn{4}{c}{SPTO (mm)} \\
    \midrule
    Model & Method & MAE & RMSE & iMAE & iRMSE & MAE & RMSE & iMAE & iRMSE & MAE & RMSE & iMAE & iRMSE \\ 
    \midrule
    \multirow{6}*{VOICED}
    & Finetuned & 3.066 & 2.475 & 4.815 & 5.571 & 64.946 & 130.229 & 17.842 & 39.909 & 51.581 & 107.829 & 15.904 & 36.604
 \\
    & EWC       & 3.189 & 1.893 & 6.874 & 8.376 & 64.850 & 129.705 & 17.955 & 40.526 & 51.639 & 108.317 & 16.055 & 37.330
 \\
    & LwF       & 8.058 & 5.496 & 11.476 & 11.741 & 68.079 & 133.513 & 20.004 & 43.943 & 53.951 & 109.904 & 17.853 & 39.579
 \\
    & Replay    & 7.588 & 4.717 & 19.605 & 20.487 & 71.119 & 139.216 & 21.754 & 46.938 & 57.968 & 116.636 & 19.699 & 42.695
\\
    & ANCL      & 14.200 & 7.096 & 20.185 & 16.408 & 71.180 & 135.583 & 19.998 & 43.718 & 53.742 & 112.189 & 17.618 & 40.213
 \\
    & CMP       & 6.923 & 4.259 & 14.240 & 18.545 & 66.639 & 132.781 & 19.699 & 44.404 & 55.596 & 115.432 & 18.129 & 41.823
 \\
    \midrule
    \multirow{6}*{FusionNet} 
    & Finetuned & -0.932 & 1.872 & -3.863 & -0.875 & 65.716 & 134.609 & 17.112 & 38.915 & 51.246 & 110.127 & 14.919 & 35.577
 \\
    & EWC       & 2.966 & 4.946 & 4.077 & 9.404 & 67.140 & 136.901 & 17.897 & 40.665 & 52.654 & 112.293 & 15.734 & 37.287
 \\
    & LwF      & -5.189 & -4.175 & -12.703 & -13.945 & 63.697 & 128.577 & 16.005 & 35.141 & 49.659 & 104.974 & 13.894 & 32.027
 \\
    & Replay    & -0.042 & 0.021 & -0.575 & -1.335 & 66.128 & 134.107 & 17.877 & 39.781 & 51.885 & 110.235 & 15.498 & 35.965
\\
    & ANCL      & 10.818 & 9.102 & 16.138 & 21.646 & 74.731 & 149.734 & 21.780 & 48.898 & 60.476 & 127.344 & 19.485 & 45.069
 \\
    & CMP       & 13.970 & 10.112 & 24.404 & 29.053 & 75.401 & 149.310 & 22.792 & 51.279 & 61.368 & 127.004 & 20.395 & 47.038
 \\
    \midrule
    \multirow{6}*{KBNet}
    & Finetuned & 16.111 & 17.678 & 9.807 & 13.391 & 63.064 & 134.674 & 15.650 & 36.275 & 44.986 & 101.266 & 12.971 & 31.393
\\
    & EWC      & 10.516 & 13.401 & 5.126 & 8.996 & 61.393 & 131.994 & 15.823 & 36.532 & 45.951 & 103.006 & 13.530 & 32.465
 \\
    & LwF       & 11.110 & 6.110 & 12.480 & 13.539 & 61.797 & 126.955 & 17.367 & 39.599 & 46.784 & 102.881 & 14.750 & 34.807
 \\
    & Replay   & 7.501 & 5.230 & 9.862 & 10.223 & 58.972 & 125.088 & 16.435 & 37.717 & 46.711 & 103.417 & 14.326 & 34.077
 \\
    & ANCL      & 6.799 & 4.797 & 8.117 & 8.647 & 61.915 & 131.275 & 17.487 & 39.455 & 49.083 & 110.329 & 15.733 & 37.085
 \\
    & CMP       & 14.005 & 9.561 & 24.587 & 29.524 & 64.330 & 135.738 & 19.138 & 44.763 & 51.741 & 115.460 & 17.089 & 41.576
 \\
    \midrule
\end{tabular}
}
\vspace{-1mm}
\label{tab:indoor2}
\end{table*}

\begin{table*}[t]
\scriptsize
\centering
\caption{\textbf{Results on outdoors}. Models are trained on KITTI and continually trained on Waymo, then VKITTI. 
}

\setlength\tabcolsep{2pt}
\resizebox{1.0\textwidth}{!}{%
\begin{tabular}{c l c c c c@{\hspace{7pt}} c c c c@{\hspace{7pt}} c c c c}
    \toprule
    & & \multicolumn{4}{c}{Average Forgetting (\%)} & \multicolumn{4}{c}{Average Performance (mm)\hspace*{6pt}} & \multicolumn{4}{c}{SPTO (mm)} \\
    \midrule
    Model & Method & MAE & RMSE & iMAE & iRMSE & MAE & RMSE & iMAE & iRMSE & MAE & RMSE & iMAE & iRMSE \\ 
    \midrule
    \multirow{6}*{VOICED}
    & Finetuned
    & 499.598 & 162.188 & 467.472 & 208.693  & 1620.429 & 3072.129 & 4.040 & 6.144& 914.223 & 2993.228 & 1.955 & 4.503
    \\
    & EWC
    & 555.925 & 190.152 & 540.109 & 247.943  & 1796.300 & 3346.057 & 4.490 & 6.685& 962.937 & 3209.759 & 1.962 & 4.739
    \\
    & LwF
    & 631.119 & 221.535 & 524.976 & 233.758  & 1973.972 & 3612.700 & 4.533 & 6.648& 985.995 & 3236.244 & 2.062 & 4.722
    \\
    & Replay
    & 17.241 & 4.050 & 16.662 & 5.478 & 524.114 & 1875.897 & 1.333 & 3.359& 618.668 & 2366.577 & 1.292 & 3.348 
    \\
    & ANCL
    & 178.694 & 64.382 & 170.133 & 80.818 & 1019.338 & 2317.409 & 2.392 & 4.425 & 881.934 & 2641.127 & 1.790 & 3.965
    \\
    & CMP
    & 37.689 & 14.711 & 70.463 & 50.115 & 1451.306 & 3226.971 & 2.319 & 4.396 & 1491.603 & 3686.281 & 2.081 & 4.200

    \\
    \midrule
    \multirow{6}*{FusionNet} 
    & Finetuned
    & 11.336 & 8.435 & 17.447 & 17.991  & 437.730 & 1785.212 & 1.193 & 3.724& 501.362 & 2138.422 & 1.111 & 3.978
    \\
    & EWC
    & 21.006 & 10.494 & 20.431 & 16.535 & 431.440 & 1760.460 & 1.144 & 3.181 & 486.170 & 2117.030 & 1.029 & 2.986
    \\
    & LwF
    & 12.368 & 5.202 & 13.593 & 13.117 & 442.878 & 1759.202 & 1.178 & 3.352 & 526.528 & 2168.961 & 1.156 & 3.451
    \\
    & Replay
    & 8.290 & 11.134 & 2.769 & 7.975  & 419.044 & 1774.361 & 1.044 & 3.032& 479.168 & 2122.997 & 0.966 & 2.906
    \\
    & ANCL
    & 6.717 & 7.089 & 6.164 & 10.803 & 404.664 & 1716.474 & 1.071 & 3.159 & 459.187 & 2055.770 & 0.988 & 3.065
    \\
    & CMP
    & 5.585 & 8.860 & 1.383 & 7.466 & 405.654 & 1743.608 & 1.042 & 3.073 & 462.593 & 2087.346 & 0.968 & 2.969
    \\
    \midrule
    \multirow{6}*{KBNet}
    & Finetuned
    & 27.153 & 18.208 & 52.969 & 33.370 & 469.658 & 1943.259 & 1.338 & 3.683& 541.383 & 2411.169 & 1.144 & 3.505 
    \\
    & EWC
    & 23.517 & 8.583 & 30.077 & 18.991  & 456.828 & 1806.761 & 1.221 & 3.321& 526.366 & 2210.424 & 1.133 & 3.158
    \\
    & LwF
    & 21.184 & 4.049 & 43.500 & 19.951& 460.097 & 1749.734 & 1.362 & 3.555 & 541.932 & 2142.999 & 1.359 & 3.731 
    \\ 
    & Replay
    & 25.423 & 29.303 & 6.362 & 7.274 & 454.896 & 1935.667 & 1.102 & 3.203& 525.696 & 2318.363 & 1.094 & 3.246 
    \\
    & ANCL
    & 20.49 & 8.94 & 23.11 & 27.73 & 438.05 & 1795.76 & 1.21 & 3.56 & 503.53 & 2203.44 & 1.18 & 3.53 
    \\
    & CMP
    & 15.95 & 15.47 & 6.90 & 7.39 & 507.90 & 2262.46 & 1.06 & 3.21 & 447.09 & 1887.14 & 1.09 & 3.19
    \\
    \midrule
\end{tabular}
}
\vspace{-5mm}
\label{tab:outdoor1}
\end{table*}

\begin{table*}[t]
\scriptsize
\centering
\vspace{-2mm}
\caption{\textbf{Results on outdoor}. Models are trained on KITTI and continually trained on VKITTI, then Waymo. 
}
\setlength\tabcolsep{2pt}
\resizebox{1.0\textwidth}{!}{%
\begin{tabular}{c l c c c c@{\hspace{7pt}} c c c c@{\hspace{7pt}} c c c c}
    \toprule
    & & \multicolumn{4}{c}{Average Forgetting (\%)} & \multicolumn{4}{c}{Average Performance (mm)\hspace*{6pt}} & \multicolumn{4}{c}{SPTO (mm)} \\
    \midrule
    Model & Method & MAE & RMSE & iMAE & iRMSE & MAE & RMSE & iMAE & iRMSE & MAE & RMSE & iMAE & iRMSE \\ 
    \midrule
    \multirow{6}*{VOICED}
    & Finetuned & 474.009 & 165.136 & 370.424 & 155.125 & 2060.582 & 4376.605 & 3.562 & 6.024 & 1025.089 & 3418.751 & 2.040 & 4.630
 \\
    & EWC       & 461.303 & 147.778 & 364.455 & 135.980 & 1750.973 & 3812.288 & 3.325 & 5.560 & 965.069 & 3216.165 & 1.968 & 4.434
\\
    & LwF       & 337.637 & 104.948 & 273.390 & 107.768 & 1494.371 & 3476.494 & 2.871 & 5.185 & 942.391 & 3125.526 & 1.923 & 4.296
 \\
    & Replay    & 13.913 & 11.382 & 27.166 & 8.929 & 584.114 & 2329.848 & 1.304 & 3.274 & 615.781 & 2396.354 & 1.263 & 3.075
 \\
    & ANCL      & 159.673 & 50.002 & 197.543 & 81.902 & 1113.680 & 2997.533 & 2.347 & 4.725 & 877.832 & 2869.267 & 1.799 & 4.107
 \\
    & CMP      & 48.249 & 11.697 & 4.621 & 4.479 & 1070.327 & 2700.213 & 1.616 & 3.672 & 1440.935 & 3237.217 & 1.949 & 3.785
 \\
    \midrule
    \multirow{6}*{FusionNet} 
    & Finetuned & 16.344 & 10.358 & 27.541 & 17.882 & 470.609 & 2082.091 & 1.093 & 3.355 & 510.531 & 2161.420 & 1.092 & 3.161
 \\
    & EWC       & 12.313 & 9.516 & 35.145 & 23.740 & 459.507 & 93.598 & 1.082 & 3.367 & 493.153 & 2109.876 & 1.077 & 3.164
 \\
    & LwF       & 1.651 & -13.290 & 37.645 & 7.821 & 436.941 & 1932.231 & 1.090 & 3.347 & 496.067 & 2239.943 & 1.042 & 3.379
 \\
    & Replay   & 6.715 & 9.282 & 6.068 & 7.796 & 436.402 & 2023.981 & 0.956 & 3.057 & 469.038 & 2082.781 & 0.966 & 2.907
 \\
    & ANCL      & 6.814 & 7.446 & 11.704 & 13.729 & 435.777 & 2007.397 & 1.019 & 3.225 & 466.817 & 2069.395 & 1.019 & 3.038
 \\
    & CMP      & 9.658 & 10.924 & 6.204 & 8.565 & 437.079 & 2030.872 & 0.988 & 3.186 & 466.370 & 2088.647 & 0.987 & 3.028
 \\
    \midrule
    \multirow{6}*{KBNet}
    & Finetuned & 19.842 & 11.959 & 17.144 & 14.162 & 469.095 & 2076.653 & 1.185 & 3.513 & 496.070 & 2140.853 & 1.155 & 3.299
 \\
    & EWC       & 18.438 & 14.418 & 32.407 & 20.567 & 471.227 & 2131.648 & 1.141 & 3.476 & 501.447 & 2191.964 & 1.132 & 3.285
\\
    & LwF       & 14.797 & 4.244 & 68.109 & 33.271 & 496.194 & 2076.227 & 1.476 & 4.028 & 535.322 & 2170.847 & 1.415 & 3.830
\\
    & Replay    & 13.929 & 17.453 & 7.582 & 6.629 & 458.919 & 2156.319 & 1.023 & 3.279 & 491.371 & 2216.334 & 1.038 & 3.147
 \\
    & ANCL      & 14.748 & 9.927 & 22.279 & 15.974 & 473.733 & 2151.693 & 1.062 & 3.419 & 505.116 & 2225.249 & 1.068 & 3.251
 \\
    & CMP       & 15.170 & 15.887 & 9.719 & 7.491 & 463.297 & 2158.487 & 1.023 & 3.297 & 497.619 & 2229.364 & 1.036 & 3.166
 \\
    \midrule
\end{tabular}
}
\vspace{-5mm}
\label{tab:outdoor2}
\end{table*}

\subsection{Experience Replay}

To perform experience replay with depth completion, we maintain a replay buffer, which retains a fixed number of representative samples from each dataset previously trained on. 
The selections and the integration of the replay buffer into the new task are designed to balance computational efficiency with performance retention. Empirically, a buffer size of 64 data points for each previous dataset was determined to offer an optimal balance. During each training iteration on a new target dataset, a batch of data from the replay buffer is reintroduced into the training process. The data points from both the current and historical datasets are processed as follows:

Let \( D_{\text{new}} = D_k \) denote the set of new training data points from the current target dataset, where \( N \) is the number of new data points. Similarly, let \( D_{\text{replay}} = \{ (I_{r}^{(j)}, z_{r}^{(j)}, K_{r}^{(j)}) \}_{j=1}^{M} \) denote the set of replayed data points from the buffer, where \( M \) is the number of data points in the replay buffer. Each training batch is made up of a 50-50 new-replay data ratio, where the replay half is evenly split between previous datasets. The total loss function \(\mathcal{L}_{\text{total}}\) for a training iteration is then formulated as:
\begin{equation}
\mathcal{L}_{\text{total}} = \mathcal{L}_{\text{new}} + \mathcal{L}_{\text{replay}}
\end{equation}
where \( \mathcal{L}_{\text{new}} \) is the unsupervised loss $\mathcal{L}$ in Eq. \ref{eqn:objective_function} computed on the new training data, and \( \mathcal{L}_{\text{replay}} \) is the unsupervised loss $\mathcal{L}$ computed on the replayed data. 

\subsubsection{Optimal Memory Buffer Retention Strategy}
As a more recent replay-based method, we implemented the strategy from the CMP~\cite{kang2024continual} method to more carefully select which samples to store in the replay buffer. Instead of randomly choosing a set of samples from the previous datasets to store in the replay buffer, we attempt to store samples that show a similarity below some threshold \(\delta\). We do this by iterating through the dataset we want to replay in the next continual step \(D_{new}\), and comparing the similarity of the next batch in this dataset \(D_{new}^{(b)}\) and a randomly sampled batch that is already stored in the replay buffer \(D_{replay}^{(b)}\) after being passed through a representation encoder \(f_{\theta}\). Thus, at each step of the iteration, we append \(D_{new}^{(b)}\) to  \(D_{replay}\) if similarity between $f_{\theta}(D_{new}^{(b)})$ and $f_{\theta}(D_{replay}^{(b)})$ is less than  $\delta$.

\subsection{Auxiliary Network Continual Learning}
Auxiliary Network Continual Learning (ANCL) introduces an auxiliary model to promote plasticity, complementing the main model’s focus on stability. For continual depth completion, we implement ANCL by maintaining two networks: the main model \( f_\theta \), and an auxiliary network \( f_{\theta_\text{aux}} \), both initialized from the previously trained model on dataset \( D_{k-1} \).

When training on a new dataset \( D_k \), both the frozen parameters \( \theta^* \) (from previous dataset) and the auxiliary model parameters \( \theta_\text{aux} \) are used to compute constraints based on stability and plasticity. Specifically, the ANCL loss is a weighted sum of two Elastic Weight Consolidation (EWC) losses:
\begin{equation}
\mathcal{L}_{\text{ancl}} = \lambda_{\text{ewc}} \sum_i \frac{1}{2} F_i (\theta_i - \theta^*_{i})^2 + \lambda_{\text{aux}} \sum_i \frac{1}{2} F_i (\theta_i - \theta_{\text{aux}, i})^2
\end{equation}
Here, the Fisher information matrix \( F_i \) is estimated for both the main and auxiliary models during continual training. To train the auxiliary model, we compute an unsupervised loss (Eq.~\ref{eqn:objective_function}) independently using the same inputs, and backpropagate this loss through \( f_{\theta_\text{aux}} \) using a separate optimizer. 
We set \( \lambda_{\text{ewc}} = 1.0 \) and \( \lambda_{\text{aux}} = 0.5 \) based on validation performance. Fisher matrices for both networks are accumulated across epochs and updated after training on each dataset, consistent with EWC.

\begin{figure*}[t]
    \centering
    \includegraphics[width=0.86\textwidth]{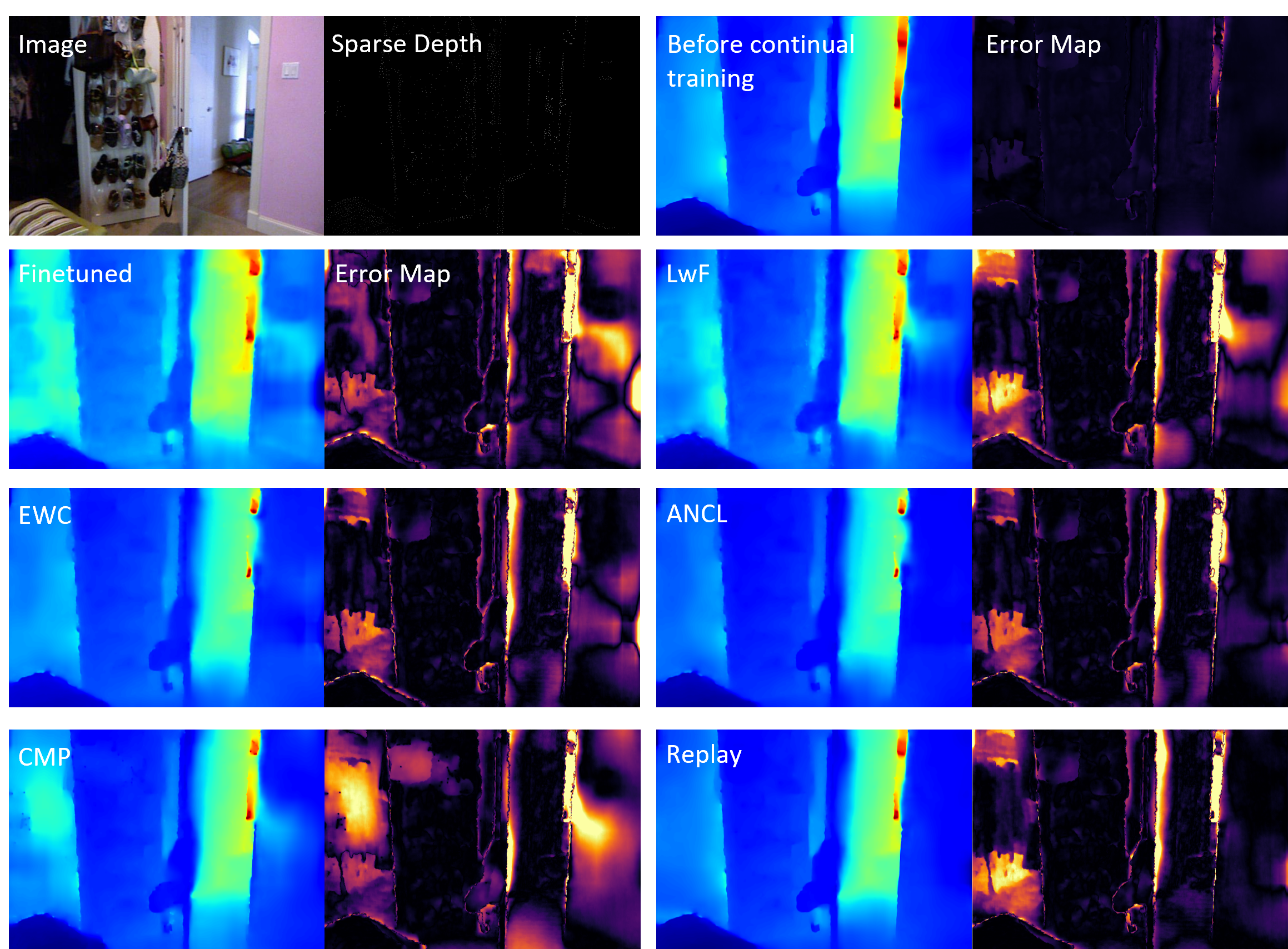}
    \vspace{-1mm}
    \caption{\textbf{Comparison on NYUv2 after continual training on VOID.}
    Rows 2-4 show results from models that were first trained on NYUv2 and then continually trained on VOID using different continual learning methods.}
    \label{fig:qualitative_comparison}
    \vspace{-6mm}
\end{figure*}

\begin{table}[t!]
\centering
\footnotesize
    \vspace{-0.8em}
    \caption{
        \textit{Error metrics.} There are \(N\) datasets and \(k\) denotes the index of the most recently trained dataset \(D_k\). Each \(a_j^k\) represents any of our depth completion metrics on dataset \(D_j\) after training on dataset \(D_k\).
    }
    \vspace{0.8em}
\renewcommand{\arraystretch}{1.3}

\begin{tabular}{l l}
    \midrule
        Metric & Definition \\ \midrule
        Average Forgetting ($\bar{F}$) &$\frac{2}{N(N-1)}\, \sum_{k=1}^N\, \sum_{j<k}\, \frac{a_j^k - a_j^j}{a_j^j}$ \\
        \addlinespace[0.25em]
        Average Performance ($\bar{\mu}$) & $ \frac{2}{N(N+1)}\, \sum_{k=1}^N\, \sum_{j\leq k}\, a_j^k$ \\
        \addlinespace[0.5em]
        Stability-Plasticity Trade-off (SPTO) & $\frac{2\times S\times P}{S+P},\,
        \begin{cases}
        S = \sum_{k=1}^N a_k^N \\[0.25em]
        P = \sum_{k=1}^N a_k^k\,
        \end{cases}$ \\ \midrule
        \rule{0pt}{2.5ex} 
    \end{tabular}
\label{tab:error_metrics}
\vspace{-10mm}
\end{table}

\section{Experiments}

For evaluation, we compute four three metrics: Average Forgetting, Average Performance, and STPO on Mean Absolute Error (MAE), Root Mean Squared Error (RMSE), Inverse MAE (iMAE), and Inverse RMSE (iRMSE). Average Forgetting measures how much performance on previously learned datasets deteriorates after learning new ones. Average Performance reflects the overall accuracy across all seen datasets. Stability-Plasticity Trade-off (SPTO) quantifies the balance between preserving prior knowledge (stability) and acquiring new knowledge (plasticity) using a harmonic mean of final and intermediate performance. These metrics are computed over \textit{all datasets}. See Table \ref{tab:error_metrics} for details.

\textbf{Datasets.}
We use three datasets for indoor experiments and three datasets for outdoor experiments. 

\textit{Indoor datasets:} The \textbf{NYUv2} dataset \cite{silberman2012indoor} contains 464 indoor scenes captured using RGB-D sensors, offering 1449 densely labeled pairs of aligned RGB and depth images. It is a benchmark dataset widely used in indoor depth estimation tasks. NYUv2 is the primary dataset on which we pretrain our depth completion models before outdoor continual learning. The \textbf{VOID} dataset \cite{wong2020unsupervised} provides sparse depth maps and RGB frames for indoor environments, with approximately 58,000 frames. VOID emphasizes handling low-texture regions and scenes with significant camera motion, which are crucial for testing robustness in indoor depth completion. \textbf{ScanNet} \cite{dai2017scannet} is a large-scale indoor dataset that includes more than 2.5 million frames with corresponding RGB-D data. It provides dense depth ground truth and 3D reconstructions of indoor environments. We use the evaluation protocol in \cite{wong2020unsupervised} and cap the evaluation range from 0.2 to 5 meters. 

\textit{Outdoor datasets: }The \textbf{KITTI} dataset \cite{geiger2012we} is a widely used benchmark for autonomous driving, consisting of over 93,000 stereo image pairs and sparse LiDAR depth maps synchronized with the images that were captured from a wider range of urban and rural scenes. KITTI is the primary dataset on which we pretrain our depth completion models before outdoor continual learning. The \textbf{Waymo} \cite{sun2020scalability} Open Dataset includes approximately 230,000 frames of high-resolution images and dense LiDAR point clouds, covering a wide range of driving environments and conditions. \textbf{Virtual KITTI (VKITTI)},  \cite{gaidon2016virtual} is a synthetic dataset designed to replicate KITTI scenes, providing over 21,000 frames with dense ground truth depth. It allows for evaluating domain adaptation, as we use multiple data domains to simulate domain discrepancies which cause catastrophic forgetting. For KITTI and VKITTI, we cap the evaluation range from 0.001 to 100 meters. For Waymo, we cap the evaluation range from 0.001 to 80 meters.


\textbf{Protocol.} For indoor sequences, we first train on NYUv2 and evaluate the continual learning process over sequences: NYUv2 $\rightarrow$ ScanNet $\rightarrow$ VOID (Table~\ref{tab:indoor1}), and NYUv2 $\rightarrow$ VOID $\rightarrow$ ScanNet (Table~\ref{tab:indoor2}).
For outdoor sequences, we first train on KITTI and evaluate over sequences:
KITTI $\rightarrow$ Waymo $\rightarrow$ Virtual KITTI (Table~\ref{tab:outdoor1}), and KITTI $\rightarrow$ Virtual KITTI $\rightarrow$ Waymo (Table~\ref{tab:outdoor2}). For the mixed-domain (indoor-outdoor) sequence, we first train on KITTI and evaluate the continual learning process over KITTI $\rightarrow$ NYUv2 $\rightarrow$ Waymo. Due to space constraints, the mixed experiments are positioned in the Supp. Mat. Figure~\ref{fig:path} illustrates all datasets and the colored arrow paths corresponding to each of these training sequences used in our benchmark. All experiments were conducted using two servers, each equipped with 4 NVIDIA RTX 3090 GPUs (24GB VRAM per GPU).

\subsection{Results}

\textbf{Quantitative.} Due to space limitations, we defer raw results of MAE, RMSE, iMAE, and iRMSE across all datasets to the Supp. Mat. From Tables~\ref{tab:indoor1} and~\ref{tab:indoor2}, we observe that for indoor continual learning, most methods outperform the finetuned baseline in terms of Average Forgetting across multiple models. In particular, \textbf{Replay} and \textbf{CMP} consistently outperform Finetuned in Table~\ref{tab:indoor1} across all metrics. \textbf{EWC} and \textbf{LwF} offer modest improvements over Finetuned in certain FusionNet and KBNet rows, but are less reliable in VOICED. Surprisingly, the finetuning approach performs the worst in terms of Average Forgetting in \textit{only 3 out of 6} scenarios (model-sequence pairing) for indoor datasets, and \textit{4 out of 6} scenarios for outdoor datasets. In Average Performance, Finetuned only performs the worst in \textit{3 out of 12} scenarios. This shows that when encountering complicated 3D tasks, existing continual learning methods are not guaranteed to improve performance. 

For outdoor datasets (Tables~\ref{tab:outdoor1} and~\ref{tab:outdoor2}), the performance gap between methods is larger. \textbf{Replay} reduces forgetting significantly compared to Finetuned in most rows, particularly in Table~\ref{tab:outdoor1} for VOICED (17.241\% vs. 499.598\% forgetting). However, in Table~\ref{tab:outdoor2}, the forgetting advantage of Replay narrows, and \textbf{CMP} achieves comparable forgetting in FusionNet and KBNet. \textbf{EWC} and \textbf{LwF} occasionally show worse forgetting than Finetuned outdoors, especially in VOICED, but achieve competitive SPTO in some FusionNet cases. \textbf{ANCL} performs stably in KBNet and FusionNet, but loses to replay-based methods in VOICED. In addition, methods with the best Average Forgetting score do not always have the best Average Performance score. For example, in Table~\ref{tab:outdoor1}, CMP has the lowest Average Forgetting with KBNet but is the worst in terms of Average Performance. 
These findings suggest that while replay-based methods often yield strong forgetting reduction, regularization-based methods ANCL can be competitive depending on the model and training sequence. Overall, \textit{no single method dominates across all metrics and settings}, highlighting the importance of evaluating forgetting, learning, and trade-offs jointly in continual depth completion. This also leaves continual learning for depth completion tasks a challenging open question. 

\textbf{Qualitative.} Fig.~\ref{fig:qualitative_comparison} shows that all continual learning methods develop artifacts in their predictions for the previously trained dataset, NYUv2 -- even though they are only adapting to the first dataset of the sequence. When compared with the original pretrained model for NYUv2, they all perform poorly as they are not able to mitigate catastrophic forgetting. Additional examples are available in Supp. Mat.


\section{Discussion}


\textbf{Limitations.} The scope of our benchmark is restricted to existing canonical continual learning methods adapted to unsupervised depth completion; newer or task-specific methods remain unexplored. Also, we leave exploration of unsupervised continual learning for other tasks as future work. 

\textbf{Societal Impacts.} Continual learning reduces the need to train separate models for new domains, lowering computational energy costs. It also promotes "backwards-compatibility" and positive-congruent models. However, it introduces the possibility for the model to forget safety-critical knowledge or learn unintended biases when exposed to uncurated or non-representative data.

To the best of our knowledge, this is the first study to apply continual learning methods to the problem of depth completion. By establishing a benchmark for continual depth completion across indoor and outdoor datasets, this paper sets the foundation for future work in extending continual learning to more complex, multimodal problems. We hope that this contribution will motivate further research into more effective strategies to combat catastrophic forgetting in depth completion and related 3D tasks. We invite the research community to join us in tackling this important open problem.

{\small
\small
\clearpage
\bibliographystyle{IEEEtran}
\bibliography{root}

\begin{thebibliography}{10}
\providecommand{\url}[1]{#1}
\csname url@rmstyle\endcsname
\providecommand{\newblock}{\relax}
\providecommand{\bibinfo}[2]{#2}
\providecommand\BIBentrySTDinterwordspacing{\spaceskip=0pt\relax}
\providecommand\BIBentryALTinterwordstretchfactor{4}
\providecommand\BIBentryALTinterwordspacing{\spaceskip=\fontdimen2\font plus
\BIBentryALTinterwordstretchfactor\fontdimen3\font minus \fontdimen4\font\relax}
\providecommand\BIBforeignlanguage[2]{{%
\expandafter\ifx\csname l@#1\endcsname\relax
\typeout{** WARNING: IEEEtran.bst: No hyphenation pattern has been}%
\typeout{** loaded for the language `#1'. Using the pattern for}%
\typeout{** the default language instead.}%
\else
\language=\csname l@#1\endcsname
\fi
#2}}

\bibitem{park2020non}
J.~Park, K.~Joo, Z.~Hu, C.~Liu, and I.-S. Kweon, ``Non-local spatial propagation network for depth completion,'' in \emph{European Conference on Computer Vision (ECCV)}, 2020, pp. 120--136.

\bibitem{kam2022costdcnet}
J.~Kam, J.~Kim, S.~Kim, J.~Park, and S.~Lee, ``Costdcnet: Cost volume based depth completion for a single rgb-d image,'' in \emph{European Conference on Computer Vision}.\hskip 1em plus 0.5em minus 0.4em\relax Springer, 2022, pp. 257--274.

\bibitem{tang2024bilateral}
J.~Tang, F.-P. Tian, B.~An, J.~Li, and P.~Tan, ``Bilateral propagation network for depth completion,'' in \emph{Proceedings of the IEEE/CVF Conference on Computer Vision and Pattern Recognition}, 2024, pp. 9763--9772.

\bibitem{wong2020unsupervised}
A.~Wong, X.~Fei, S.~Tsuei, and S.~Soatto, ``Unsupervised depth completion from visual inertial odometry,'' \emph{IEEE Robotics and Automation Letters}, vol.~5, no.~2, pp. 1899--1906, 2020.

\bibitem{wong2021unsupervised}
A.~Wong and S.~Soatto, ``Unsupervised depth completion with calibrated backprojection layers,'' in \emph{Proceedings of the IEEE/CVF International Conference on Computer Vision}, 2021, pp. 12\,747--12\,756.

\bibitem{yan2023desnet}
Z.~Yan, K.~Wang, X.~Li, Z.~Zhang, J.~Li, and J.~Yang, ``Desnet: Decomposed scale-consistent network for unsupervised depth completion,'' in \emph{Proceedings of the AAAI Conference on Artificial Intelligence}, ser. IEEE Robotics and Automation Letters, 2023, pp. 3109--3117.

\bibitem{geiger2012we}
A.~Geiger, P.~Lenz, and R.~Urtasun, ``Are we ready for autonomous driving? the kitti vision benchmark suite,'' in \emph{2012 IEEE Conference on Computer Vision and Pattern Recognition}.\hskip 1em plus 0.5em minus 0.4em\relax IEEE, 2012, pp. 3354--3361.

\bibitem{frahm2010building}
J.-M. Frahm, P.~Fite-Georgel, D.~Gallup, T.~Johnson, R.~Raguram, C.~Wu, Y.-H. Jen, E.~Dunn, B.~Clipp, S.~Lazebnik, \emph{et~al.}, ``Building rome on a cloudless day,'' in \emph{Computer Vision--ECCV 2010: 11th European Conference on Computer Vision, Heraklion, Crete, Greece, September 5-11, 2010, Proceedings, Part IV 11}.\hskip 1em plus 0.5em minus 0.4em\relax Springer, 2010, pp. 368--381.

\bibitem{schonberger2016structure}
J.~L. Schonberger and J.-M. Frahm, ``Structure-from-motion revisited,'' in \emph{Proceedings of the IEEE conference on computer vision and pattern recognition}, 2016, pp. 4104--4113.

\bibitem{french1999catastrophic}
R.~M. French, ``Catastrophic forgetting in connectionist networks,'' \emph{Trends in cognitive sciences}, vol.~3, no.~4, pp. 128--135, 1999.

\bibitem{mccloskey1989catastrophic}
M.~McCloskey and N.~J. Cohen, ``Catastrophic interference in connectionist networks: The sequential learning problem,'' in \emph{Psychology of learning and motivation}.\hskip 1em plus 0.5em minus 0.4em\relax Elsevier, 1989, vol.~24, pp. 109--165.

\bibitem{thrun1995learning}
S.~Thrun, ``Is learning the n-th thing any easier than learning the first?'' \emph{Advances in neural information processing systems}, vol.~8, 1995.

\bibitem{ratcliff1990connectionist}
R.~Ratcliff, ``Connectionist models of recognition memory: constraints imposed by learning and forgetting functions.'' \emph{Psychological review}, vol.~97, no.~2, p. 285, 1990.

\bibitem{yan2021positive}
S.~Yan, Y.~Xiong, K.~Kundu, S.~Yang, S.~Deng, M.~Wang, W.~Xia, and S.~Soatto, ``Positive-congruent training: Towards regression-free model updates,'' in \emph{Proceedings of the IEEE/CVF Conference on Computer Vision and Pattern Recognition}, 2021, pp. 14\,299--14\,308.

\bibitem{wong2021learning}
A.~Wong, S.~Cicek, and S.~Soatto, ``Learning topology from synthetic data for unsupervised depth completion,'' \emph{IEEE Robotics and Automation Letters}, vol.~6, no.~2, pp. 1495--1502, 2021.

\bibitem{kirkpatrick2017overcoming}
J.~Kirkpatrick, R.~Pascanu, N.~Rabinowitz, J.~Veness, G.~Desjardins, \emph{et~al.}, ``Overcoming catastrophic forgetting in neural networks,'' \emph{Proceedings of the national academy of sciences}, vol. 114, no.~13, pp. 3521--3526, 2017.

\bibitem{li2017learning}
Z.~Li and D.~Hoiem, ``Learning without forgetting,'' in \emph{Proceedings of the IEEE conference on computer vision and pattern recognition}, 2017, pp. 5077--5086.

\bibitem{kim2022ancl}
S.~Kim, L.~Noci, A.~Orvieto, and T.~Hofmann, ``Achieving a better stability-plasticity trade-off via auxiliary networks in continual learning,'' in \emph{Proceedings of the IEEE/CVF Conference on Computer Vision and Pattern Recognition (CVPR)}, 06 2023, pp. 11\,930--11\,939.

\bibitem{rolnick2019experience}
D.~Rolnick, A.~Ahuja, J.~Schwarz, T.~Lillicrap, and G.~Wayne, ``Experience replay for continual learning,'' \emph{Advances in neural information processing systems}, vol.~32, 2019.

\bibitem{kang2024continual}
D.~Kang, D.~Kum, and S.~Kim, ``Continual learning for motion prediction model via meta-representation learning and optimal memory buffer retention strategy,'' in \emph{Proceedings of the IEEE/CVF Conference on Computer Vision and Pattern Recognition}, 2024, pp. 15\,438--15\,448.

\bibitem{silberman2012indoor}
N.~Silberman, D.~Hoiem, P.~Kohli, and R.~Fergus, ``Indoor segmentation and support inference from rgbd images,'' in \emph{European conference on computer vision}.\hskip 1em plus 0.5em minus 0.4em\relax Springer, 2012, pp. 746--760.

\bibitem{dai2017scannet}
A.~Dai, A.~X. Chang, M.~Savva, M.~Halber, T.~Funkhouser, and M.~Nie{\ss}ner, ``Scannet: Richly-annotated 3d reconstructions of indoor scenes,'' in \emph{Proceedings of the IEEE conference on computer vision and pattern recognition}, 2017, pp. 5828--5839.

\bibitem{sun2020scalability}
P.~Sun, H.~Kretzschmar, X.~Dotiwalla, A.~Chouard, V.~Patnaik, P.~Tsui, J.~Guo, Y.~Zhou, Y.~Chai, B.~Caine, \emph{et~al.}, ``Scalability in perception for autonomous driving: Waymo open dataset,'' in \emph{Proceedings of the IEEE/CVF Conference on Computer Vision and Pattern Recognition}, 2020, pp. 2446--2454.

\bibitem{gaidon2016virtual}
A.~Gaidon, Q.~Wang, Y.~Cabon, and E.~Vig, ``Virtual kitti: An annotated virtual dataset for scene understanding,'' in \emph{Proceedings of the IEEE conference on computer vision and pattern recognition workshops}, 2016, pp. 28--37.

\bibitem{zenke2017continual}
F.~Zenke, B.~Poole, and S.~Ganguli, ``Continual learning through synaptic intelligence,'' in \emph{International conference on machine learning}.\hskip 1em plus 0.5em minus 0.4em\relax PMLR, 2017, pp. 3987--3995.

\bibitem{aljundi2018memory}
R.~Aljundi, F.~Babiloni, M.~Elhoseiny, M.~Rohrbach, and T.~Tuytelaars, ``Memory aware synapses: Learning what (not) to forget,'' in \emph{Proceedings of the European conference on computer vision (ECCV)}, 2018, pp. 139--154.

\bibitem{chaudhry2018riemannian}
A.~Chaudhry, P.~K. Dokania, T.~Ajanthan, and P.~H. Torr, ``Riemannian walk for incremental learning: Understanding forgetting and intransigence,'' in \emph{Proceedings of the European conference on computer vision (ECCV)}, 2018, pp. 532--547.

\bibitem{hinton2015distilling}
G.~Hinton, O.~Vinyals, and J.~Dean, ``Distilling the knowledge in a neural network,'' \emph{arXiv preprint arXiv:1503.02531}, 2015.

\bibitem{dhar2019learning}
P.~Dhar, R.~V. Singh, K.-C. Peng, Z.~Wu, and R.~Chellappa, ``Learning without memorizing,'' in \emph{Proceedings of the IEEE/CVF conference on computer vision and pattern recognition}, 2019, pp. 5138--5146.

\bibitem{rannen2017encoder}
A.~Rannen, R.~Aljundi, M.~B. Blaschko, and T.~Tuytelaars, ``Encoder based lifelong learning,'' in \emph{Proceedings of the IEEE international conference on computer vision}, 2017, pp. 1320--1328.

\bibitem{lee2019overcoming}
K.~Lee, K.~Lee, J.~Shin, and H.~Lee, ``Overcoming catastrophic forgetting with unlabeled data in the wild,'' in \emph{Proceedings of the IEEE/CVF International Conference on Computer Vision}, 2019, pp. 312--321.

\bibitem{douillard2020podnet}
A.~Douillard, M.~Cord, C.~Ollion, T.~Robert, and E.~Valle, ``Podnet: Pooled outputs distillation for small-tasks incremental learning,'' in \emph{Proceedings of the IEEE/CVF Conference on Computer Vision and Pattern Recognition (CVPR)}, 2020, pp. 1951--1960.

\bibitem{hou2019learning}
S.~Hou, X.~Pan, C.~Change~Loy, Z.~Wang, and D.~Lin, ``Learning a unified classifier incrementally via rebalancing,'' in \emph{Proceedings of the IEEE/CVF Conference on Computer Vision and Pattern Recognition (CVPR)}, 2019, pp. 831--839.

\bibitem{titsias2019functional}
M.~K. Titsias, J.~Schwarz, A.~G. d.~G. Matthews, R.~Pascanu, and Y.~W. Teh, ``Functional regularisation for continual learning with gaussian processes,'' \emph{arXiv preprint arXiv:1901.11356}, 2019.

\bibitem{pan2020continual}
Y.~Pan, F.~Li, and W.~K. Lee, ``Continual deep learning by functional regularisation of memorable past,'' \emph{arXiv preprint arXiv:2007.15302}, 2020.

\bibitem{nguyen2018variational}
C.~V. Nguyen, Y.~Li, T.~D. Bui, and R.~E. Turner, ``Variational continual learning,'' in \emph{International Conference on Learning Representations}, 2018.

\bibitem{chaudhry2019tiny}
A.~Chaudhry, M.~Rohrbach, M.~Elhoseiny, T.~Ajanthan, P.~K. Dokania, P.~H. Torr, and M.~Ranzato, ``On tiny episodic memories in continual learning,'' \emph{arXiv preprint arXiv:1902.10486}, 2019.

\bibitem{riemer2018learning}
M.~Riemer, I.~Cases, R.~Ajemian, M.~Liu, I.~Rish, Y.~Tu, and G.~Tesauro, ``Learning to learn without forgetting by maximizing transfer and minimizing interference,'' \emph{arXiv preprint arXiv:1810.11910}, 2018.

\bibitem{vitter1985random}
J.~S. Vitter, ``Random sampling with a reservoir,'' \emph{ACM Transactions on Mathematical Software (TOMS)}, vol.~11, no.~1, pp. 37--57, 1985.

\bibitem{lopez2017gradient}
D.~Lopez-Paz and M.~Ranzato, ``Gradient episodic memory for continual learning,'' \emph{Advances in neural information processing systems}, vol.~30, 2017.

\bibitem{rebuffi2017icarl}
S.-A. Rebuffi, A.~Kolesnikov, G.~Sperl, and C.~H. Lampert, ``icarl: Incremental classifier and representation learning,'' in \emph{Proceedings of the IEEE conference on Computer Vision and Pattern Recognition}, 2017, pp. 2001--2010.

\bibitem{shin2017continual}
H.~Shin, J.~K. Lee, J.~Kim, and J.~Kim, ``Continual learning with deep generative replay,'' \emph{Advances in neural information processing systems}, vol.~30, 2017.

\bibitem{riemer2019scalable}
M.~Riemer, T.~Klinger, D.~Bouneffouf, and M.~Franceschini, ``Scalable recollections for continual lifelong learning,'' in \emph{Proceedings of the AAAI conference on artificial intelligence}, 2019, pp. 1352--1359.

\bibitem{rostami2019complementary}
M.~Rostami, S.~Kolouri, and P.~K. Pilly, ``Complementary learning for overcoming catastrophic forgetting using experience replay,'' \emph{arXiv preprint arXiv:1903.04566}, 2019.

\bibitem{pfulb2021continual}
B.~Pf{\"u}lb, A.~Gepperth, and B.~Bagus, ``Continual learning with fully probabilistic models,'' \emph{arXiv preprint arXiv:2104.09240}, 2021.

\bibitem{kemker2017fearnet}
R.~Kemker and C.~Kanan, ``Fearnet: Brain-inspired model for incremental learning,'' \emph{arXiv preprint arXiv:1711.10563}, 2017.

\bibitem{gopalakrishnan2022knowledge}
S.~Gopalakrishnan, P.~R. Singh, H.~Fayek, S.~Ramasamy, and A.~Ambikapathi, ``Knowledge capture and replay for continual learning,'' in \emph{Proceedings of the IEEE/CVF winter conference on applications of computer vision}, 2022, pp. 10--18.

\bibitem{ayub2021eec}
A.~Ayub and A.~R. Wagner, ``Eec: Learning to encode and regenerate images for continual learning,'' \emph{arXiv preprint arXiv:2101.04904}, 2021.

\bibitem{ostapenko2019learning}
O.~Ostapenko, M.~Puscas, T.~Klein, P.~Jahnichen, and M.~Nabi, ``Learning to remember: A synaptic plasticity driven framework for continual learning,'' in \emph{Proceedings of the IEEE/CVF conference on computer vision and pattern recognition}, 2019, pp. 11\,321--11\,329.

\bibitem{wu2018memory}
Y.~X. Wu, L.~Herranz, X.~Liu, J.~van~de Weijer, B.~Raducanu, and T.~Tuytelaars, ``Memory replay gans: Learning to generate new categories without forgetting,'' in \emph{Proceedings of the 32nd International Conference on Neural Information Processing Systems}, 2018, pp. 5967--5977.

\bibitem{zhu2022self}
K.~Zhu, W.~Zhai, Y.~Cao, J.~Luo, and Z.-J. Zha, ``Self-sustaining representation expansion for non-exemplar class-incremental learning,'' in \emph{Proceedings of the IEEE/CVF Conference on Computer Vision and Pattern Recognition}, 2022, pp. 9296--9305.

\bibitem{liu2020generative}
X.~Liu, C.~Wu, M.~Menta, L.~Herranz, B.~Raducanu, A.~D. Bagdanov, S.~Jui, and J.~v. de~Weijer, ``Generative feature replay for class-incremental learning,'' in \emph{Proceedings of the IEEE/CVF Conference on Computer Vision and Pattern Recognition Workshops}, 2020, pp. 226--227.

\bibitem{iscen2020memory}
A.~Iscen, J.~Zhang, S.~Lazebnik, and C.~Schmid, ``Memory-efficient incremental learning through feature adaptation,'' in \emph{Computer Vision--ECCV 2020: 16th European Conference, Glasgow, UK, August 23--28, 2020, Proceedings, Part XVI 16}.\hskip 1em plus 0.5em minus 0.4em\relax Springer, 2020, pp. 699--715.

\bibitem{belouadah2019il2m}
E.~Belouadah and A.~Popescu, ``Il2m: Class incremental learning with dual memory,'' in \emph{Proceedings of the IEEE/CVF international conference on computer vision}, 2019, pp. 583--592.

\bibitem{chawla2024continual}
H.~Chawla, A.~Varma, E.~Arani, and B.~Zonooz, ``Continual learning of unsupervised monocular depth from videos,'' in \emph{Proceedings of the IEEE/CVF Winter Conference on Applications of Computer Vision}, 2024, pp. 8419--8429.

\bibitem{vodisch2023codeps}
N.~V{\"o}disch, K.~Petek, W.~Burgard, and A.~Valada, ``Codeps: Online continual learning for depth estimation and panoptic segmentation,'' \emph{arXiv preprint arXiv:2303.10147}, 2023.

\bibitem{yang2019dense}
Y.~Yang, A.~Wong, and S.~Soatto, ``Dense depth posterior (ddp) from single image and sparse range,'' in \emph{Proceedings of the IEEE/CVF Conference on Computer Vision and Pattern Recognition}, 2019, pp. 3353--3362.

\bibitem{liu2022monitored}
T.~Y. Liu, P.~Agrawal, A.~Chen, B.-W. Hong, and A.~Wong, ``Monitored distillation for positive congruent depth completion,'' in \emph{Computer Vision--ECCV 2022: 17th European Conference, Tel Aviv, Israel, October 23--27, 2022, Proceedings, Part II}.\hskip 1em plus 0.5em minus 0.4em\relax Springer, 2022, pp. 35--53.

\bibitem{wu2024augundo}
Y.~Wu, T.~Y. Liu, H.~Park, S.~Soatto, D.~Lao, and A.~Wong, ``Augundo: Scaling up augmentations for monocular depth completion and estimation,'' in \emph{European Conference on Computer Vision}.\hskip 1em plus 0.5em minus 0.4em\relax Springer, 2024.

\bibitem{park2024test}
H.~Park, A.~Gupta, and A.~Wong, ``Test-time adaptation for depth completion,'' in \emph{Proceedings of the IEEE/CVF Conference on Computer Vision and Pattern Recognition}, 2024, pp. 20\,519--20\,529.

\bibitem{hu2021penet}
M.~Hu, S.~Wang, B.~Li, S.~Ning, L.~Fan, and X.~Gong, ``Penet: Towards precise and efficient image guided depth completion,'' in \emph{2021 IEEE International Conference on Robotics and Automation (ICRA)}.\hskip 1em plus 0.5em minus 0.4em\relax IEEE, 2021, pp. 13\,656--13\,662.

\bibitem{li2020multi}
A.~Li, Z.~Yuan, Y.~Ling, W.~Chi, C.~Zhang, \emph{et~al.}, ``A multi-scale guided cascade hourglass network for depth completion,'' in \emph{Proceedings of the IEEE/CVF Winter Conference on Applications of Computer Vision}, 2020, pp. 32--40.

\bibitem{eldesokey2020uncertainty}
A.~Eldesokey, M.~Felsberg, K.~Holmquist, and M.~Persson, ``Uncertainty-aware cnns for depth completion: Uncertainty from beginning to end,'' in \emph{Proceedings of the IEEE/CVF Conference on Computer Vision and Pattern Recognition (CVPR)}, 2020, pp. 12\,014--12\,023.

\bibitem{van2019sparse}
W.~Van~Gansbeke, D.~Neven, B.~De~Brabandere, and L.~Van~Gool, ``Sparse and noisy lidar completion with rgb guidance and uncertainty,'' in \emph{2019 16th International Conference on Machine Vision Applications (MVA)}.\hskip 1em plus 0.5em minus 0.4em\relax IEEE, 2019, pp. 1--6.

\bibitem{zhang2018deep}
Y.~Zhang and T.~Funkhouser, ``Deep depth completion of a single rgb-d image,'' in \emph{Proceedings of the IEEE Conference on Computer Vision and Pattern Recognition}, 2018, pp. 175--185.

\bibitem{zhang2023completionformer}
Y.~Zhang, X.~Guo, M.~Poggi, Z.~Zhu, G.~Huang, and S.~Mattoccia, ``Completionformer: Depth completion with convolutions and vision transformers,'' in \emph{Proceedings of the IEEE/CVF Conference on Computer Vision and Pattern Recognition}, 2023, pp. 18\,527--18\,536.

\bibitem{yu2023aggregating}
Z.~Yu, Z.~Sheng, Z.~Zhou, L.~Luo, S.~Cao, H.~Gu, H.~Zhang, and H.~Shen, ``Aggregating feature point cloud for depth completion,'' in \emph{Proceedings of the IEEE/CVF International Conference on Computer Vision}, 2023, pp. 8732--8743.

\bibitem{ma2019self}
F.~Ma, G.~Cavalheiro, and S.~Karaman, ``Self-supervised sparse-to-dense: Self-supervised depth completion from lidar and monocular camera,'' in \emph{2019 International Conference on Robotics and Automation (ICRA)}.\hskip 1em plus 0.5em minus 0.4em\relax IEEE, 2019, pp. 3288--3295.

\bibitem{shivakumar2019dfusenet}
S.~S. Shivakumar, T.~Nguyen, I.~D. Miller, S.~W. Chen, V.~Kumar, and C.~J. Taylor, ``Dfusenet: Deep fusion of rgb and sparse depth information for image guided dense depth completion,'' in \emph{2019 IEEE Intelligent Transportation Systems Conference (ITSC)}.\hskip 1em plus 0.5em minus 0.4em\relax IEEE, 2019, pp. 13--20.

\bibitem{wong2021adaptive}
A.~Wong, X.~Fei, B.-W. Hong, and S.~Soatto, ``An adaptive framework for learning unsupervised depth completion,'' \emph{IEEE Robotics and Automation Letters}, vol.~6, no.~2, pp. 3120--3127, 2021.

\bibitem{lepetit2009epnp}
V.~Lepetit, F.~Moreno-Noguer, and P.~Fua, ``Epnp: An accurate o (n) solution to the pnp problem,'' \emph{International journal of computer vision}, vol.~81, no.~2, p. 155, 2009.

\bibitem{fischler1981random}
M.~A. Fischler and R.~C. Bolles, ``Random sample consensus: a paradigm for model fitting with applications to image analysis and automated cartography,'' \emph{Communications of the ACM}, vol.~24, no.~6, pp. 381--395, 1981.

\bibitem{lopez2020project}
A.~Lopez-Rodriguez, B.~Busam, and K.~Mikolajczyk, ``Project to adapt: Domain adaptation for depth completion from noisy and sparse sensor data,'' in \emph{Proceedings of the Asian Conference on Computer Vision (ACCV)}, 2020.

\bibitem{jeon2022struct}
J.~Jeon, H.~Lim, D.~U. Seo, and H.~Myung, ``Struct-mdc: Mesh-refined unsupervised depth completion leveraging structural regularities from visual slam,'' in \emph{IEEE Robotics and Automation Letters (RA-L)}, no.~3, 2022, pp. 6391--6398.

\end{thebibliography}
}

\end{document}